%% file: emnlp2021.tex
\documentclass[11pt]{article}

\usepackage[]{emnlp2021}

\usepackage{times}
\usepackage{latexsym}

\usepackage[T1]{fontenc}

\usepackage[utf8]{inputenc}

\usepackage{microtype}

\usepackage{booktabs}
\usepackage{multirow}
\usepackage{graphicx}
\usepackage{epstopdf}
\usepackage{array}
\usepackage{pgfplots}
\usepackage{wrapfig}

\usepackage{amsmath}

\usepackage{pifont} 
\newcommand{\cmark}{\ding{51}}%
\newcommand{\xmark}{\ding{55}}%

\usepackage{url}
\usepackage{xcolor}
\usepackage{setspace}

\input{acl-ijcnlp2021-templates/commands}

\makeatletter
\newcommand{\printfnsymbol}[1]{%
  \textsuperscript{\@fnsymbol{#1}}%
}
\makeatother

\title{\dataset: Numerical Reasoning with Interpretable Graph Question Answering Dataset}

\author{
    Qiyuan Zhang$^{1}$\thanks{$\;\;$The first three authors contributed equally. The order of authorship is decided through dice rolling.} 
    ~~Lei Wang$^{2}$\printfnsymbol{1}
   ~~Sicheng Yu$^{2}$\printfnsymbol{1}
   ~~Shuohang Wang \\ \bf
   ~~Yang Wang$^{3}$ 
   ~~Jing Jiang$^{2}$
   ~~Ee-Peng Lim$^{2}$\\ 
   $^{1}$University of Electronic Science and Technology of China \\
   $^{2}$Singapore Management University $^{3}$ Verily Life Sciences\\
  \small \texttt{qiyuanzhang97@gmail.com,}~~
  \small \texttt{\{lei.wang.2019,scyu.2018\}@phdcs.smu.edu.sg,}\\
  \small \texttt{utleonwang@gmail.com,}~~
  \small \texttt{\{shwang.2014,jingjiang,eplim\}@smu.edu.sg}
}

\begin{document}

\maketitle
\input{acl-ijcnlp2021-templates/section/0_abs}

\input{acl-ijcnlp2021-templates/section/1_intro}

\input{acl-ijcnlp2021-templates/section/2_task_definition}

\input{acl-ijcnlp2021-templates/section/3_data_collection}

\input{acl-ijcnlp2021-templates/section/4_data_analysis}

\input{acl-ijcnlp2021-templates/section/5_model}

\input{acl-ijcnlp2021-templates/section/6_experiment}
\input{acl-ijcnlp2021-templates/section/7_related_work}

\input{acl-ijcnlp2021-templates/section/8_conclusion}
\bibliography{emnlp2021}
\bibliographystyle{acl_natbib}

\clearpage

\input{acl-ijcnlp2021-templates/9_appendix}

\end{document}

%% file: acl-ijcnlp2021-templates/commands.tex
\newcommand{\ie}{\textit{i.e.}}
\newcommand{\eg}{\textit{e.g.}}

\newcommand{\dataset}{\textsc{Noah}QA}
\newcommand{\model}{RGNet}

%% file: acl-ijcnlp2021-templates/section/0_abs.tex
\begin{abstract}
While diverse question answering (QA) datasets have been proposed and contributed significantly to the development of deep learning models for QA tasks, the existing datasets fall short in two aspects.
First, we lack QA datasets covering complex questions that 
involve answers as well as the reasoning processes to get the answers. As a result, the state-of-the-art QA research on numerical reasoning still focuses on simple calculations and does not provide the mathematical expressions or evidences justifying the answers.
Second, the QA community has contributed much effort to improving the interpretability of QA models.
However, these models fail to explicitly show the reasoning process, such as the evidence order for reasoning and the interactions between different pieces of evidence.
To address the above shortcomings, we introduce \dataset{}, a conversational and bilingual QA dataset with questions requiring numerical reasoning with compound mathematical expressions.
With \dataset{}, we develop an interpretable reasoning graph as well as the appropriate evaluation metric to measure the answer quality. 
We evaluate the state-of-the-art QA models trained using existing QA datasets on \dataset{} and show that the best among them can only achieve 55.5 exact match scores, while the human performance is 89.7. 
We also present a new QA model for generating a reasoning graph where the reasoning graph metric still has a large gap compared with that of humans, \eg, 28 scores. The dataset and code are publicly available  \footnote{\url{https://github.com/Don-Joey/NoahQA}}.

\end{abstract}

%% file: acl-ijcnlp2021-templates/section/1_intro.tex
\section{Introduction}
Question answering (QA) plays a core role in natural language understanding and is a proxy to evaluate the reading comprehension ability of intelligent systems.
\input{acl-ijcnlp2021-templates/image/tisser_figure}
Due to its profound significance, a surge of datasets, \eg, span-extraction~\citep{rajpurkar2016squad}, multiple-choice~\citep{lai2017race} and open-domain~\citep{kwiatkowski2019natural}, have been proposed recently. However, those datasets have limitations in numerical reasoning and interpretability, which hinder the further advancement of QA community.


On the one hand, numerical reasoning is one of the intelligent skills of human beings. To endow such an ability to QA models, we need to provide some math word question answering dataset~\cite{Ling2017ProgramIB} for training these QA models. In particular, given a math question, a good QA model should select the correct answer among the multiple pre-defined answer options.
Recently, DropQA~\cite{dua2019drop} includes numerical questions into conventional span-extraction question answering. However, the mathematical forms of numerical questions found in DropQA are relatively simple, \eg, most of the questions are only about addition and subtraction. Besides, DropQA only provides the final answer without a full expression to the answer.

\input{acl-ijcnlp2021-templates/image/reasoning_compare}

On the other hand, existing QA datasets reveal drawbacks in interpretability (or explainability).
As the leaderboards of QA datasets are overwhelmed by powerful language models, \eg, BERT~\cite{devlin2019bert}, researchers turn their attention to explainable model and the collection of QA datasets likewise pursue this trend: models should not only give the answers, but also the explanation for the answers, \eg, complex questions with discrete reasoning or numerical reasoning. The model with explanation is also friendly with digging and improving the system. To achieve the goal,
CoQA~\cite{reddy2019coqa} and HotpotQA~\cite{yang2018hotpotqa} provide the model with rational and evidences as additional supervisions. Nonetheless, how the model conducts reasoning is still vague. For example, how the model process several pieces of evidence? Are they processed in parallel or follow a specific order? $\mathrm{R^4C}$~\cite{inoue2020r4c}, \
and 2WikiMultiHopQA~\cite{ho2020constructing} solve this problem to some extent by introducing a set of triplets or reasoning path, which is not suitable in the scenario where the reasoning process is complicated. Taking the example in Figure~\ref{fig:example} for illustration, the answer of $Q_7$ comes from the answer of $Q_5$, and the answer of $Q_5$ comes from the passage and the answer of $Q_3$.

\input{acl-ijcnlp2021-templates/table/data_comparison}
To address the above shortcomings of existing QA datasets, we present \dataset{}, Numerical reasOning with interpretAble grapH QA dataset. An overall comparison of the differences between \dataset{} and others is shown in Table~\ref{tab:data_comparison}. \dataset{} is constructed in a conversational and bilingual form with fruitful complex numerical questions demanding addition, subtraction, multiplication, division, and combination of parentheses.
Meanwhile, 
\dataset{} provides annotated explanations in the form of reasoning graphs, namely a graph of reasoning steps, to explicitly represent the global reasoning process for each question.
The comparison of reasoning graph and other explanation annotations is shown in Figure~\ref{fig:explain_compare}.
Reasoning graph can be used as supervision in training as well as surrogate for evaluating the interpretability of QA models.

We apply strong baselines from existing datasets on \dataset{} and discover that the best baseline achieves $55.50$ exact match scores, while human performance is $89.67$.
For evaluating of reasoning graph, we introduce an automatic evaluation method named $\mathrm{DAG}_{sim}$, considering the structural and semantic similarity at the same time between the predicted and ground-truth reasoning graphs.
To facilitate the research along with reasoning graph, we also contribute a new model named Reasoning Graph Network (\model) for generating the reasoning graph. Experiments with \model{} show that there is still a large gap behind human performance with $28.01$ and $18.14$ $\mathrm{DAG}_{sim}$ scores on the English and Chinese versions of \dataset{}, respectively. 


%% file: acl-ijcnlp2021-templates/image/tisser_figure.tex
\begin{figure}[h!]
 \centering
    \includegraphics[width=0.44\textwidth]{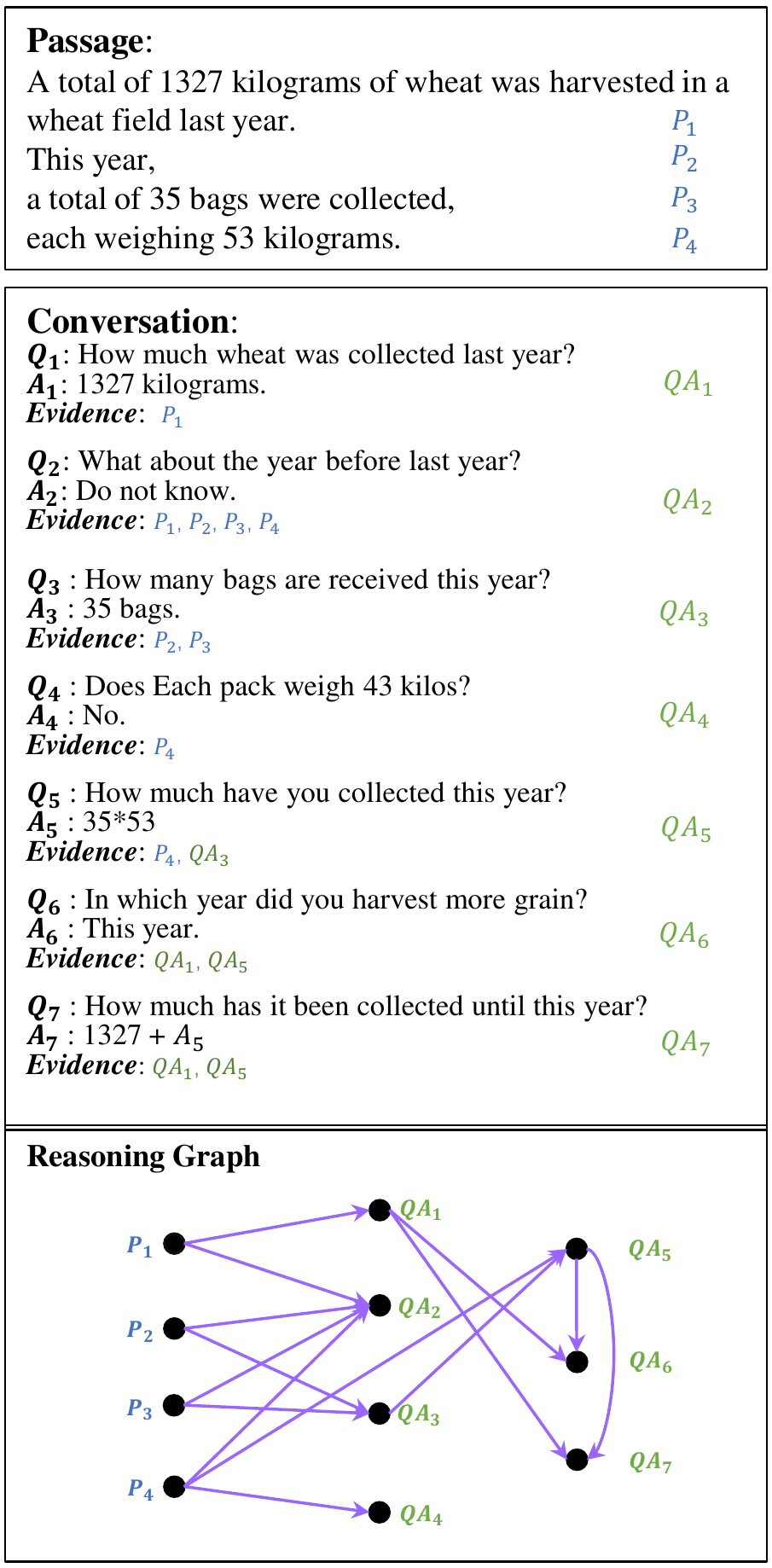}
    \caption{A sample of \dataset{} dataset, which consists of a passage and several question-answer pairs. The supporting evidences and reasoning graph are provided for correctness and interpretability evaluation.}
    \vspace{-0.56cm}
    \label{fig:example}
\end{figure}

%% file: acl-ijcnlp2021-templates/image/reasoning_compare.tex
\begin{wrapfigure}{l}{0.25\textwidth}
    \centering
    \includegraphics[width=0.23\textwidth]{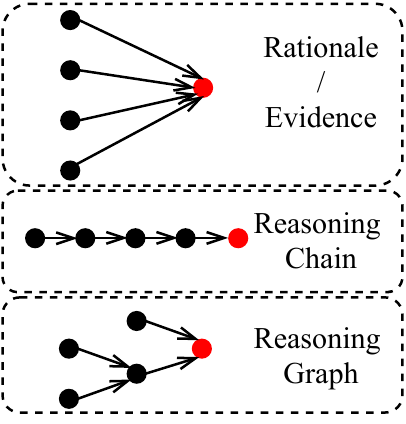}
    \caption{Comparison between rationale/evidence (HotpotQA and CoQA), reasoning chain (2WikiMultiHopQA) and  reasoning graph (\dataset).}
    \label{fig:explain_compare}
\end{wrapfigure}

%% file: acl-ijcnlp2021-templates/table/data_comparison.tex
\begin{table*}[t]
    \centering
    \resizebox{0.99\textwidth}{!}{
    \begin{tabular}{ l c c c c c l p{9.2cm}}
    \toprule
    Dataset & Conversational & Cross-lingual &  Mathematics& Expression & Evidence & Explanation Annotations   \\ \midrule
    CoQA~\cite{reddy2019coqa} & \cmark & \xmark &\xmark &\xmark &\cmark & Rationale (text span) \\
    HotpotQA~\cite{yang2018hotpotqa} & \xmark & \xmark & \xmark & \xmark & \cmark & Evidence (set of supporting facts)\\
    $\mathrm{R}^4\mathrm{C}$~\cite{inoue2020r4c}&\xmark  & \xmark & \xmark &\xmark &\cmark & Derivation (set of triplets)\\
    2WikiMultiHopQA~\cite{ho2020constructing}&\xmark  & \xmark & \xmark &\xmark &\cmark & Reasoning Path (chain of triplets)\\
    
    DropQA~\cite{dua2019drop}&\xmark  & \xmark & \cmark &\xmark &\xmark & - \\
    
    Dream~\cite{sun2019dream}&\xmark  & \xmark & \cmark &\xmark &\xmark &-\\
    
    MathQA~\cite{amini-etal-2019-mathqa}&\xmark  & \xmark & \cmark &\cmark &\xmark & Operation Programs\\
    
    Math23K~\cite{wang-etal-2017-deep}&\xmark  & \xmark & \cmark &\cmark &\xmark & -\\

    \midrule
    \dataset & \cmark & \cmark &\cmark &\cmark & \cmark & Reasoning Graph \\
    \bottomrule
    \end{tabular}%
    }
    \caption{Comparison of \dataset{} with existing datasets. One may argue that expression equals to evidence which is not always the case, \eg, both of ``two years'' and ``two miles'' contain ``two'' but with totally different semantic meanings.
    }
    \vspace{-10px}
    \label{tab:data_comparison}
\end{table*}

%% file: acl-ijcnlp2021-templates/section/2_task_definition.tex
\section{Task}
\label{sec:task}
Following previous works on math word questions~\cite{amini-etal-2019-mathqa}, our first task is still to generate answers for all questions in conversations. Besides, in order to get an explainable model, our second task is to generate the reasoning graph when answering questions. Formally, given a background passage $P$, a history conversation $QA_{1:(t-1)}$ and the next question $Q_t$, the task is to return a textual answer $A_t$ to the next question $Q_t$ and generate a reasoning graph $\hat{G}^r_t$.
Next we will introduce the detailed notations.

\noindent\textbf{Textual Answer.}
Each sample in our dataset consists of a background passage $P$ splitted into a sequence of segments $\{P_1, P_2, ..., P_n\}$ by punctuations, and a conversation $QA_{1:(t-1)}$  with a series of question-answer pairs $\{QA_1, QA_2, ..., QA_{t-1}\}$.
Each answer $A_i$ ($1 \leq i \leq t$) is associated with a set of first-order evidences, $E_i$, which can be some text segments in $P$ and/or first-order evidences of previous question-answer pairs. These first-order evidences provide the information to derive the answer $A_t$ for $Q_t$ directly. Unlike most QA datasets~\cite{rajpurkar2016squad}, we only use exact match (EM) score as our evaluation metric for answer correctness, the F1 score is not adopted to bypass the occasion that two different but overlapped numbers may give high F1 scores, \eg, $1203.4$ and $1204.4$.

\noindent\textbf{Reasoning Graph.}
We now define the reasoning graph (RG) for question $Q_t$ to be a directed acyclic graph (DAG) $G^r_t=\langle \mathcal{V}_t, \mathcal{L}_t\rangle$.
For example in Figure~\ref{fig:example}, $\mathcal{V}_t$ denotes the set of first order evidences required to derive $A_t$ and $\mathcal{L}^r_t$ denotes the set of directed edges between evidences from $\mathcal{V}_t$.  
For any next question $Q_t$, we treat it as the root node and apply the breadth-first search (BFS) to construct the RG. Specifically, BFS starts from the root node and visit all of the neighbor nodes, \ie, the first-order evidence. If the present evidence is leaf node (segment in $P$), BFS stops. 
Otherwise, BFS will continue to explore its first-order evidence.
Formally, for $Q_t$ we denote the ground-truth and predicted RG as $G_t$ and $\hat{G}_t$ respectively.

To evaluate the quality of predicted reasoning graph $\hat{G}_t$,
we propose $\mathrm{DAG}_{sim}$, an automatic evaluation method considering the structural and semantic similarity between two DAGs.
We first decompose the ground-truth graph $G_t$ and the predicted graph $\hat{G}_t$ into two sets of paths $\mathcal{P}$ and $\mathcal{\hat{P}}$, respectively. A path consists of nodes from the root node to a leaf node. Then, we compute the matrix of the best alignment scores $S\in\mathcal{R}^{|P|\times|\hat{P}|}$. Then, we use a bipartite matching algorithm over $S$ to find the optimal matching set $\Pi^*$ between $\mathcal{P}$ and $\mathcal{\hat{P}}$. $\mathrm{DAG}_{sim}$ is defined as follows:
\begin{equation}
    \mathrm{DAG}_{sim} = \sum_{\pi^*\in\Pi^*} w_{\pi^*} s_{\pi^*},
\end{equation}

where $s_{\pi^*}$ is the score from $S$ for a pair of matching paths. $w_{\pi^*}$ is a weight for the matching computed in terms of node frequency in the longer of two paths.
The best alignment score between $p_i\in \mathcal{P}$ and $p_j\in \hat{\mathcal{P}}$ can be computed as below:
\begin{equation}
  c ( p_i, p_j) = \max_{A_k \in \mathcal{A}(p_i, p_j)} c (p_i, p_j,A_k),
\end{equation}
where $\mathcal{A}(p_i, p_j)$ denotes all possible one-to-one alignments between $p_i$ and $p_j$ that do not violate chronological order. An alignment score is calculated as follows:
\begin{equation}
      c (p_i, p_j,A_k) = \sum_{(v_{i,k},v_{j,k} )\in A_k} a(v_{i,k},v_{j,k}) ,
\end{equation}

where $a(v_{i,k},v_{j,k})$ is a semantic similarity score based on the text of the two nodes.

Most of traditional graph similarity methods, \eg, Graph Edit Distance (GED), are hard to scale and ignore the semantic similarity between two nodes. Precision-recall of the expected edges ignores the semantic similarity of two text nodes. The proposed $\mathrm{DAG}_{sim}$ is more comprehensive and can run efficiently and consider structured and semantic similarity simultaneously.
 

%% file: acl-ijcnlp2021-templates/section/3_data_collection.tex
\section{Data Collection}

In this section, we describe the collection process of \dataset~which consists of raw passages preparation, conversation collection, evidence labeling (translation) and validation. We elaborate each step as follows.

\noindent\textbf{Raw Passages Preparation.}
A math word problem (MWP) is composed of a short passage and a question that naturally requires the model to carefully understand the passage and take a few steps to solve. We collected questions from two classical open-source MWP datasets: the Chinese word problem dataset Math23K~\cite{wang-etal-2017-deep} and the English word problem dataset MAWPS~\cite{koncel-kedziorski-etal-2016-mawps}.  
Then we curated $19,098$ MWPs from Math23K and $2,249$ MWPs from MAWPS as the starting points for conversations.

\noindent\textbf{Conversation Collection.}
We first hired undergraduates to create conversation collection, where each of them was provided with the annotation guidelines and examples. 
Finally, we chose qualified crowd-workers among them to complete the work. The guidelines are summarized as follows.
\underline{\textit{Conversation}}: we require annotators to provide at least five conversation turns for each passage, except for the very few short articles. 
Written conversations should be concise and natural as in real occasion, \eg, if two consecutive questions share the same subject, the subject of the latter question can be omitted.
\underline{\textit{Question-Answer pair}}: we define six types of question-answer pairs (\ie, ``Extraction'', ``Numerical Reasoning'', ``Counterfactual'', ``Comparison'', ``Yes/No'', and ``Unanswerable'') and provide annotators examples of these types
so as to encourage them to create diverse questions.
\underline{\textit{Multi-step Reasoning}}: 
to come up with questions with multi-step reasoning, we adopt questions in original MWPs as the reference to guide annotators to create conversations. 
It helps to enhance the coherence and build relationship between QA pairs in the conversation.
\underline{\textit{Answer}}: the annotated answer should be either a minimum span of the text, an equation, or a fixed content, \eg, "Yes / No / Do not know".
During the whole conversation generation process, we adopted the quality control mechanism of quantitative sampling, \ie, 
for every 100 samples, we checked 20 randomly selected ones. 
Once any error was detected within the samples, we requested the corresponding annotator to review his 100 samples of conversation.

\noindent\textbf{Evidence Labeling and Translation.}
We recruited another group of undergraduates to label the evidence for question. These annotators started with the first turn and labeled the spans from the passage or question-answer pairs in the previous turns as evidence, which directly supports the answer to the current question. The rule for labelling is that: when labeling the evidence for the $QA_{t}$, (1) if $QA_{1:(t-1)}$ provide all information, we directly label the history turn(s) that is useful in reasoning this current question; (2) if $QA_{1:(t-1)}$ cannot offer information to answer this question, and there are another sentences in the passage that can provide this information, add the text segment $P_j$ (even if $P_j$ has already been used in the conversation history).  This process is repeated until the evidence of the last turn is labeled. During evidence labeling, annotators marked the question-answer pair as incorrect once they find that the answer is not correct. 
Meanwhile, we translated the passages and annotated conversations from Chinese to English using Google Translation.

\noindent\textbf{Error Correction.}
We invited a group of English Translation major graduates to review and correct grammatical and pragmatic errors in the translated text. This is necessary as we discovered that Google translation failed to provide high-quality translation for mathematics-related text.
The common translation errors are listed in the Appendix.
In the end, we have two annotators verifying 300 randomly selected samples
and marking them as valid or invalid
for both the English and Chinese test sets. Their Cohen’s kappa scores are 88.72 and 79.83, respectively.

\noindent\textbf{Annotation De-biasing}
As suggested in previous papers \cite{clark-etal-2019-dont,kaushik2018much}, the existing benchmarks on question answering have annotation biases, which makes designing models unnecessary. We discuss different biases and our counter-measures as follows.
\underline{\textit{Annotation Style Bias}}: personal language style 
may affect conversation collection and evidence labeling. To prevent the dataset from simple repetitive style bias, we have 23 annotators involved in conversation collection, 3 in evidence labeling and 4 in error correction.
\underline{\textit{Question Bias}}:
when generating questions, annotators may prefer simple questions (e.g., Extraction) over difficult ones (e.g., Numerical Reasoning). We, therefore, use thresholds to restrict the proportions of different types of questions.
\underline{\textit{Reasoning Bias}}: 
to prevent the conversation from a complete step-by-step reasoning of numerical problem, we slightly relaxed the scope of the generated questions by inserting ``Counterfactual'', ``Comparison'', and ``Unanswerable'' (c.f. Table~\ref{tab:qa_type}).
The above problems will be examined and corrected by the mechanism of quantitative sampling mentioned before. 


%% file: acl-ijcnlp2021-templates/section/4_data_analysis.tex
\input{acl-ijcnlp2021-templates/table/statistics}

\section{Data Analysis}
For simplicity and clarity, our data analysis is based on the English version of our dataset. \dataset{} contains $21,347$ examples, each of which consists of a passage, a conversation, and corresponding evidence set. we randomly split the dataset into training, development, and test sets. The detailed statistics of the dataset are shown in Table 2. In the following, we quantitatively analyze the properties of questions, answers, and evidence in the  \dataset~dataset.

\begin{figure}[!t]
 \centering
    \includegraphics[width=0.44\textwidth]{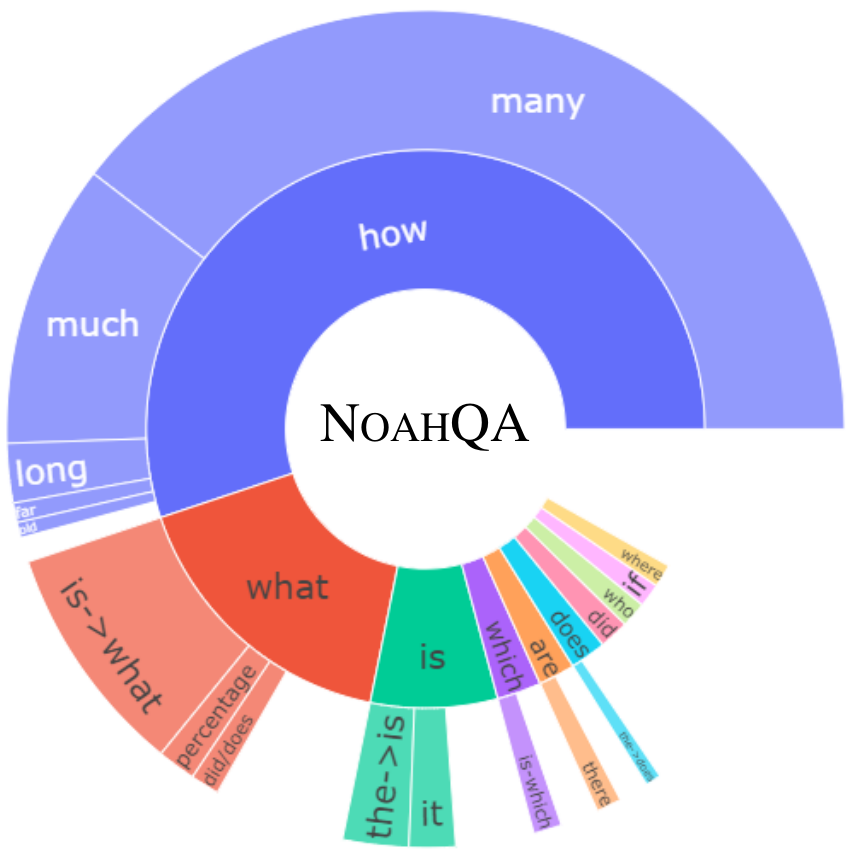}
    \caption{Distribution of question prefixes (bigram) in \dataset.}
    \vspace{-5px}
    \label{fig:question_type}
\end{figure}

\noindent\textbf{Question Analysis.}
We analyze the question types in conversation created by annotators and visualize the distribution of question types in Figure~\ref{fig:question_type}. As shown, questions beginning with ``how'' account for the vast majority, among them the questions asking about specific number are popular, \eg, ``how many'' and ``how much''. This is due to \dataset~having ``Extraction'' and ``Numerical Reasoning'' as the two most frequent QA types. 
The number of questions beginning with ``what'' and ``is'' are ranked second and third which mainly corresponds to ``Yes/No'' and ``Unanswerable'' QA types.

\input{acl-ijcnlp2021-templates/table/distribution}
\noindent\textbf{Answer Analysis.}
Based on the annotated types of answers provided by annotators, we analyze \dataset{} to assess the distribution of the answers. As shown in Table 3, Most of the answers that can be extracted in passage directly, accounting for $46.90\%$. 
Second most of the answers are numeric values ($26.22\%$) which require inferring the correct arithmetic expressions consisting of operations.
Some of the answers in this type need external knowledge. For example, people who have never learned geometric knowledge do not know how to use $\pi$ to solve circle-related problems.
Beyond, we try to mix a few counterfactual questions ($1.29\%$), where the conditions are changed, in the conversations to increase the difficulty of answering. 
The rest of the answers include ``Yes/No'' ($13.76\%$), ``Unanswerable'' ($6.47\%$), and ``Comparison'' ($5.36\%$).

\begin{figure}[!t]
 \centering
    \includegraphics[width=0.34\textwidth]{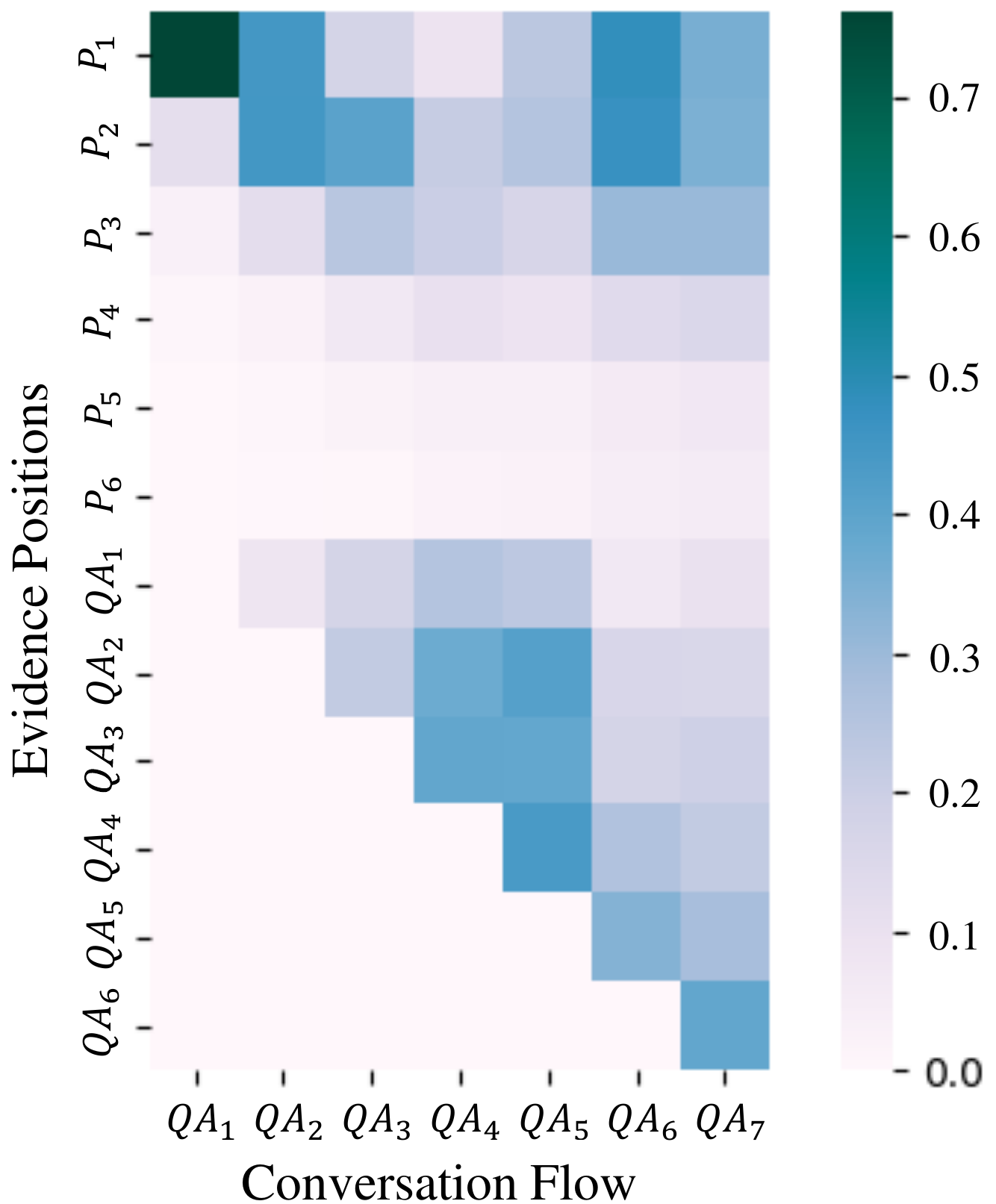}
    \caption{Distribution of first-order evidence. X-axis: the advance of the conversation, Y-axis: the evidence positions in segments and QA turns, The darker the position is, the more likely it is to be first-order evidence for the current question.}
    \vspace{-5px}
    \label{fig:evidence_analysis}
\end{figure}

\noindent\textbf{Evidence Analysis.}
Figure~\ref{fig:evidence_analysis} shows the distribution of first-order evidence of answering questions as the conversation progresses before the 8$^{th}$ turn. The first three turns of questions are prone to finding relevant first-order evidence from the first few segments of the paragraph. The 4$^{th}$ and 5$^{th}$ questions tend to capture useful information from QA pairs close to them. The 6$^{th}$ and 7$^{th}$ questions not only utilize nearby QA pairs, but also refer to the passage.  Interestingly, throughout the conversation, the segments involved in first-order evidence of all questions are the first four segments, very few questions refer to the later segments.

%% file: acl-ijcnlp2021-templates/table/statistics.tex
\begin{table*}[!t]
    \centering
    \resizebox{0.8\textwidth}{!}{
    \begin{tabular}{l|ccc|c}
    \toprule
         & Train & Dev & Test & All \\ 
    \midrule
         \# Examples & 17,077 & 2,135 & 2,135 & 21,347 \\
          Ave. / Max. \# QA-Pairs / Example & 5.08 / 10 & 5.09 / 10 & 5.09 / 9 & 5.08 / 10 \\
          Ave. / Max. \# Segments / Passage & 2.90 / 16 & 
          2.91 / 13 & 2.87 / 9 & 2.90 / 16 \\
          Ave. / Max. \# Tokens / Passage & 37.02 / 140 & 36.92 / 118 & 36.68 / 102 & 36.98 / 140 \\
          Ave. / Max. \# Tokens / Question & 8.78 / 48 & 8.77 / 40 & 8.74 / 39& 8.77 / 48 \\
          Ave. / Max. \# Tokens / Answer & 1.57 / 28 & 1.57 / 21 & 1.58 / 24& 1.57 / 28\\
          Ave. / Max. \# Evidences / Question & 2.88 / 19 & 2.86 / 15 & 2.86 / 12 & 2.88 / 19 \\
          
    \bottomrule
    \end{tabular} }
    \caption{Statistics of training, development, and test sets of \dataset.}
    \vspace{-5px}
    \label{tab:statistic}
\end{table*}

%% file: acl-ijcnlp2021-templates/table/distribution.tex
\begin{table}[t]
    \centering
    \resizebox{0.38\textwidth}{!}{
    \begin{tabular}{llp{5.2cm}}
    \toprule
        Answer Type &  Percentage  \\ \midrule
        
        Extraction & $46.90\%$  \\ \midrule

       Numerical Reasoning & $26.22\%$  \\
        \,\,\,\,w/o external knowledge & $ 25.73\%$\\ 
         \,\,\,\,w/ external knowledge& $0.49\%$ \\
        
        Counterfactual & $1.29\%$ \\  \midrule
         
        Yes/No & 13.76\% \\  \midrule
        
         Unanswerable  & $6.47\%$  \\  \midrule
        
        Comparison & 5.36\%  \\

    \bottomrule
    \end{tabular} }
    \caption{Distribution of question-answer types in \dataset. }
    \vspace{-5px}
    \label{tab:qa_type}
\end{table}

%% file: acl-ijcnlp2021-templates/section/5_model.tex
\section{\model}

\begin{figure}[t!]
  \centering
     \includegraphics[width=0.46\textwidth]{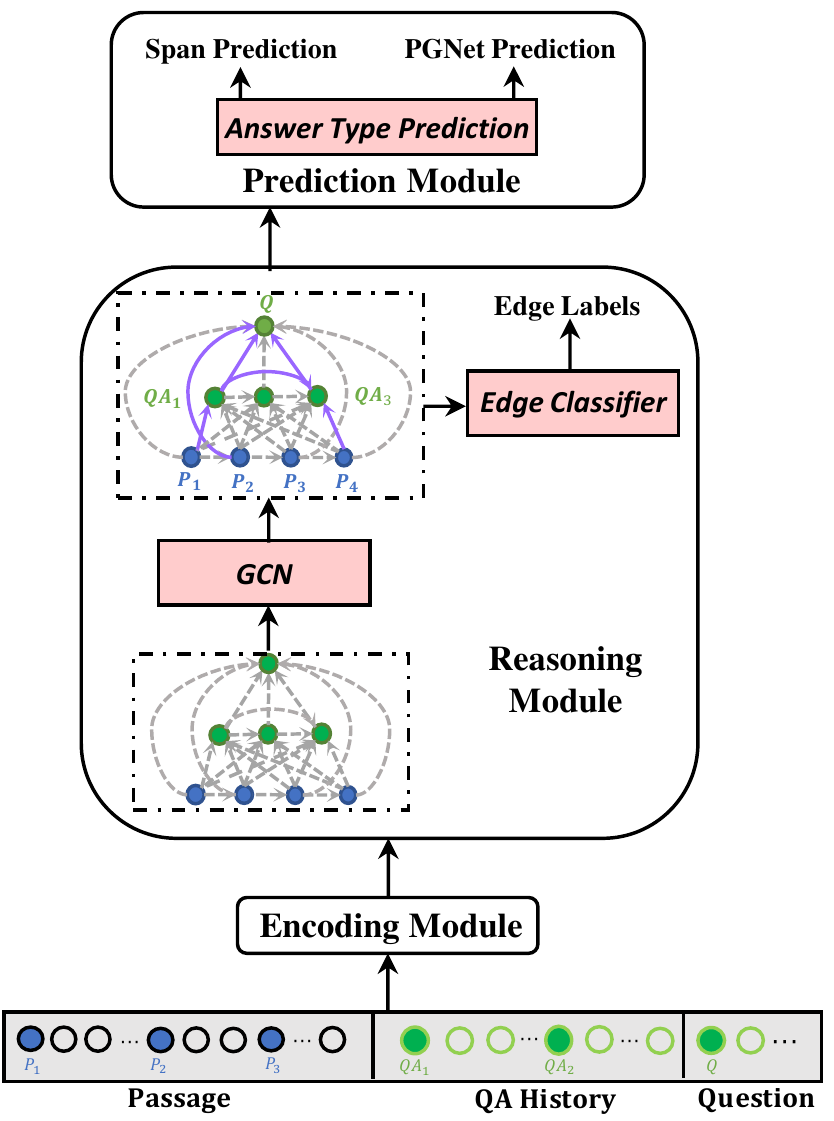}
     \caption{Framework of \model. Our model consists of an encoding module, a reasoning Module, and a prediction module.
     }
     \vspace{-5px}
     \label{fig:framework}
\end{figure}

\dataset{} requires the model to predict answers of different types, \ie, a span or an arithmetic expression, and the corresponding reasonoing graph (RG).
Here we propose Reasoning Graph Network (\model) as our baseline model. Our framework consists of three main components: an encoding module, a reasoning module, and a prediction module. The overall structure of \model{} is shown in Figure~\ref{fig:framework}.


\noindent\textbf{Encoding Module.} The input includes two channels: segments in the  passage and ground-truth history QA pairs followed by the current question.
\footnote{We add special tokens: \textit{Yes}, \textit{No}, \textit{Unknown}, and operations, \ie, \{$+,-,\times,\div$\}, to the end of the input text to facilitate the prediction for corresponding types of questions.}
We use the encoding module in NAQANet~\cite{yu2018qanet, dua2019drop} to get corresponding contextual representations.

\noindent\textbf{Reasoning Module.}
We first construct a candidate graph with the all potential edges for RG. 
Then graph convolutional network (GCN)~\cite{kipf2016semi} with $M=3$ GCN layers is utilized.
We denote two sets of representations for nodes (identifiers) and edges at $m^{th}$ GCN layer as $\mathbf{H}_n^m$ and $\mathbf{H}_n^m$, respectively.
Specifically, $\mathbf{H}_n^0$ is extracted from the output of encoding module and $\mathbf{H}_e^0$ is initialized in the beginning of training.
Given the representations at $m^{th}$ GCN layer, representations at ${(m\!+\!1)}^{th}$ GCN layer can be obtained by:
\begin{gather}
\mathbf{H}_n^{m+1} = \mathrm{Node\_Update}({G^\prime}; \mathbf{H}_n^{m} ,\mathbf{H}_e^{m}), \\
\mathbf{H}_e^{m+1} = \mathrm{Edge\_Update}({G^\prime}; \mathbf{H}_n^{m}, \mathbf{H}_e^{m}),
\end{gather}
Then we apply an edge classifier over the output of the last GCN layer, $\mathbf{H}_e^{M}$ to determine which edge in candidate graph $G^\prime$ is existed in RG.
Then we feed the combination of $\mathbf{H}_n^{M}$ and contextual representations (from encoding module) into the prediction module.  

\noindent\textbf{Prediction Module.}
For simplicity, we group six types of answer into two sets: extractive span or generative sequence.  Following NAQANet~\cite{dua2019drop}, we use an answer type prediction layer to decide the type of the answer. 
For extractive span type, we follow the standard implementation~\cite{wang2016machine} to find the start and end positions. 
For generative squence, we adopt the pointer-generator network (PGNet)~\cite{See2017GetTT}. 

\noindent\textbf{Training.}
In \model, there are four objectives during training, \ie, answer type prediction, span prediction, sequence generation, edge classification. Thus, the final loss function is defined as:
\begin{equation}
        \textit{Loss} =
    \text{XE}_{type}+
    \text{XE}_{span} + \text{XE}_{seq} + \text{XE}_{edge},
\end{equation}
where XE denotes cross-entropy loss.

%% file: acl-ijcnlp2021-templates/section/6_experiment.tex
\section{Experiment}
\subsection{Experimental Settings}
\noindent\textbf{Baselines.}
We evaluate \dataset{} on the models designed for four widely used datasets:
1) baselines from CoQA includes seq2seq implemented by OpenNMT~\cite{klein-etal-2017-opennmt}, PGNet~\cite{see2017get}, and FlowQA~\cite{huang2018flowqa}. 
These methods utilize ground-truth history answers to answer the current question; 
2) baselines  (denoted as HOTPOT)~\cite{yang2018hotpotqa} introduced in HotpotQA dataset, which utilizes the supervised signals of evidence while answering the questions; 
3) GTS~\cite{ijcai2019-736} from the MWP dataset Math23K. GTS translates an MWP text into an arithmetic expression through a goal-driven tree decoder; 
4)  NAQANet~\cite{dua2019drop} and its strong variant NumNet+~\cite{ran2019numnet} from DropQA. NAQANet can predict multiple types of answers. NumNet+ combines NAQANet with a numeric aware graph neural network. 
To establish human performance, we randomly sample 300 examples from the test set. For each example, we average evaluation scores of predictions from two annotators. 

\noindent\textbf{Implementation Details.}
In \model{}, we utilize  RoBERTa~\cite{liu2019roberta} and XLM-R~\cite{conneau-etal-2020-unsupervised} as encoding module for monolingual and cross-lingual experiments, respectively. 
Adam~\cite{kingma2014adam} is selected as optimizer in training. 
In the prediction module, we use one layer GRU~\cite{cho-etal-2014-learning} for PGNet.

\subsection{Results and Discussion}

\input{acl-ijcnlp2021-templates/table/baseline}

\noindent\textbf{Main Results.}
The results of different models on our \dataset~dataset are shown in Table~\ref{tab:baselines}. We use EM scores mentioned in Section~\ref{sec:task} as the evaluation metric for answer correctness. We observe that GTS as a strong baseline in solving MWP performs worst because it can not generate text or extract spans from the text. PGNet outperforms seq2seq due to its stronger ability to produce tokens in the paragraph and historical context. Notably,  FlowQA not only outperforms PGNet and seq2seq but also beat all of the other baselines due to its special design for conversation. HOTPOT using intermediate evidence as supervision is comparable to PGNet. NumNet+ consistently outperforms NAQANet on two versions of \dataset{} due to the numeric comparison graph.
\footnote{More experiment results under cross-lingual setting are in appendix.} 
Since \model{} is able to cover all types of questions, including GCN for reasoning, span prediction for locating text span, and PGNet for producing arithmetic expressions, it obtains overall highest performance. We also perform the one-sample t-test, and p-value = 2e-6 < 0.05 indicates that the improvements of RGNet w/o Edge over FlowQA (the strongest baseline) are statistically significant. 

From the ablative settings for \model{}, we discover that performance gain from pretraining
is considerable.
Meanwhile, GCN is also effective w/ or w/o supervision of ground-truth reasoning graph (edge classifier).
Comparing full model and the model w/o Edge, we find that edge supervision contributes to the final performance, \eg, $1.57$ points on English test set. 
From an overall view, the best model performance from \model{} is still 27.98 points and 29.82 points behind humans on the English and Chinese test sets, respectively. 

\input{acl-ijcnlp2021-templates/table/reasoning_metric}

\input{acl-ijcnlp2021-templates/table/alltypes}
\noindent\textbf{Reasoning Graph Evaluation.}
Table~\ref{tab:reasoning_graph} presents the graph exact match (GEM) and similarity score ($\mathrm{DAG}_{sim}$) for the predicted reasoning graphs where GEM measures whether two graphs are the same. 

We can clearly observe that $\mathrm{DAG}_{sim}$ is higher than GEM as it considers both structural and semantic similarities. Our model can achieve $49.21$ GEM points and $69.83$ $\mathrm{DAG}_{sim}$ scores for the English test set and $59.32$ GEM points and $79.56$ $\mathrm{DAG}_{sim}$ scores for the Chinese test set. 
 It also suggests that there is still a lot of room for improvement in generating reliable reasoning graph. 

\noindent\textbf{Analysis of Different Answer Types.}
As shown in Table~\ref{tab:answer_types}, we perform a break-down analysis of different answer types to various methods on the Chinese test set. 
GTS obtains the best result of predicting the arithmetic answers. FlowQA achieves the best results on text spans prediction.
As seq2seq and PGNet can generate arithmetic expressions, they perform better on numerical questions. 

\begin{figure}[!t]
  \centering
     \includegraphics[width=0.49\textwidth]{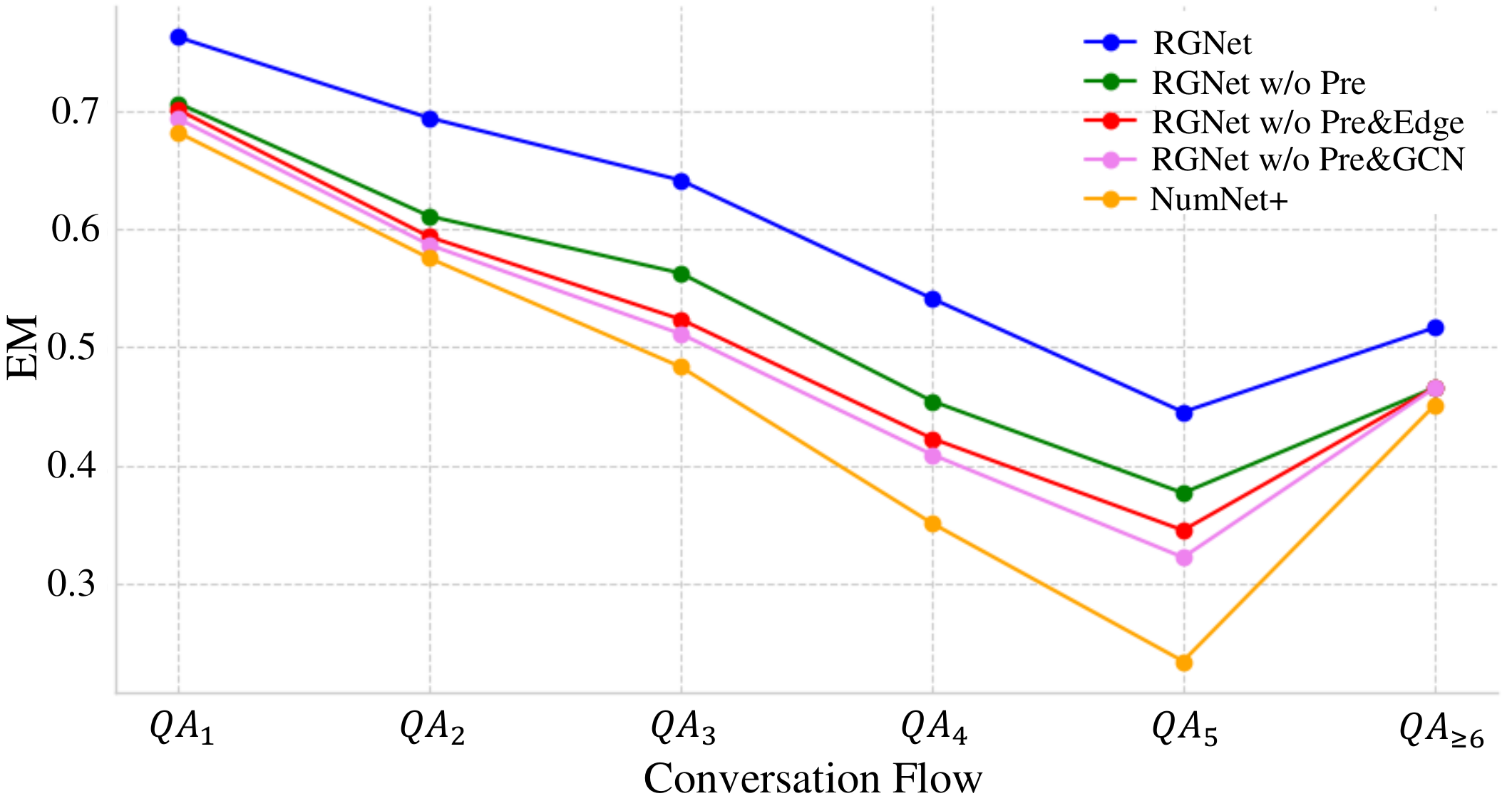}
     \caption{Performance changes in conversation flow on the test set (zh).
     }
     \vspace{-5px}
     \label{fig:performance_flow}
\end{figure}

\noindent\textbf{Analysis of Conversation Flow.}
Figure~\ref{fig:performance_flow} shows how the performance changes as the conversations progress. As the complexity of questions is increasing along the X-axis, the performance is also decreasing.
Since more unanswerable questions appear after the 6$^{th}$ question, the performances increase.
Notably, the performance of \model{} degrades more slowly than NumNet+.
\input{acl-ijcnlp2021-templates/table/cheatable}

\noindent\textbf{Analysis of the Necessity of Passages and Historical QA Pairs.}
\citet{min2019compositional} has shown that HOTPOTQA dataset has artifacts to cheat models with superficial patterns and facts in multihop reasoning are redundant. Thus, to verify that evidence is necessary in \dataset, we perform experiments on \model{} w/ or w/o passages and historical QA pairs. As shown in Table~\ref{tab:cheatable}, the significant drop in the results of model w/o passages or historical QA pairs indicates that these content are indispensable in our dataset.

\input{acl-ijcnlp2021-templates/table/transfer}

\noindent\textbf{Zero-shot Transfer.}
We perform a zero-shot transfer experiment to investigate how different our \dataset~from the complex QA dataset DropQA and the MWP dataset Math23K. 
We train NumNet+ model on DropQA and GTS model on Math23k and then test them on the English and Chinese test set of \dataset, respectively. As shown in Table~\ref{tab:transfer}, both models perform poorly, indicating that our \dataset~is vastly different from DropQA and Math23K.

%% file: acl-ijcnlp2021-templates/table/baseline.tex
\begin{table}[!t]
\footnotesize
\centering
\begin{tabular}{l | c c | c c }
\toprule
 \multirow{3}{*}{} & \multicolumn{2}{c|}{en (EM)} & \multicolumn{2}{c}{zh (EM)} \\


& Dev & Test & Dev & Test\\

\midrule
Seq2seq & 44.81& 44.93&  47.32 & 46.78 \\
PGNet & 47.33 & 47.15&  49.85 & 49.37 \\

FlowQA & 56.78 & 55.50&  55.26 & 54.02 \\ 
 
 HOTPOT & 51.24 & 51.18 &  48.02 & 47.92 \\ 
 GTS & 8.73 & 8.36 &  11.17 & 10.40 \\
 NAQANet & 51.35  & 50.45 &  47.34 & 45.78 \\
 NumNet+ & 53.04 & 52.34 &  49.60 &47.82 \\
 \midrule
 \model & \textbf{63.04}  & \textbf{61.69} &  \textbf{64.90} & \textbf{62.94} \\
 w/o Pre & 57.48  & 56.23 &  57.60 & 54.32  \\
 w/o Edge & 62.67 & 60.12 & 64.23 & 62.07 \\     
 w/o Edge\&Pre & 57.54  & 55.59 & 57.11 & 55.89 \\      
 w/o GCN  & 60.03 & 58.74 & 62.23 & 60.89 \\     
 w/o GCN\&Pre & 56.56  & 55.13 &  55.68 & 52.78 \\
\midrule
Human & -& 89.67& -& 92.76\\

\bottomrule
\end{tabular}
\caption{The experimental results of different models, including an ablation study for different RGNet variants. Pre = Pre-Trained Model (RoBERTa), Edge = Edge Classifier, GCN = Graph Convolutional Network.}
\vspace{-10px}
\label{tab:baselines}
\end{table}

%% file: acl-ijcnlp2021-templates/table/reasoning_metric.tex
\begin{table}[!t]
\footnotesize
\centering
\begin{tabular}{l | c c | c c}
\toprule
 \multirow{2}{*}{} &  \multicolumn{2}{c|}{en} & \multicolumn{2}{c}{zh} \\

  & GEM & $\mathrm{DAG}_{sim}$ & GEM & $\mathrm{DAG}_{sim}$  \\

\midrule

Edge w/o Pre &  43.14 & 63.17  &  47.15  & 66.92 \\
Edge w/ Pre & 49.21 & 69.83  & 59.32  & 79.56   \\
Human  & 93.80 & 97.84 & 92.53 & 97.71  \\



\bottomrule
\end{tabular}
\caption{Performance of \model{} w.r.t GEM score and $\mathrm{DAG}_{sim}$ score for reasoning graphs on test set.}
\label{tab:reasoning_graph}
\end{table}

%% file: acl-ijcnlp2021-templates/table/alltypes.tex
\begin{table*}[ht]
\footnotesize
\centering
\resizebox{0.97\textwidth}{!}{
\begin{tabular}{l | c| c| c| c| c| c| c| c| c}
\toprule


&~Seq2seq&~~PGNet~~&~FlowQA~&~Hotpot~&~~~GTS~~~&~NAQANet~&~NumNet+&\model* &\model \\
\midrule
Extract & 60.19 &  61.21 & 79.21&68.96& -  &  63.58 & 63.61 &65.34 & 73.20\\ 
Yes/No & 78.34 & 78.64& 81.23&76.86 &  - &  81.83 & 84.21 &85.11& 90.96\\ 
Comparison & 12.15 & 14.76&38.07&18.66&  -   &  3.37 & 26.65 &31.24& 46.56\\ 
Arithmetic & 19.45  & 23.34&-&-&  37.80  &  1.02 & 1.16 &24.12& 33.73\\ 
Counterfactual  & 20.12 & 24.51  &11.06& 2.14 & -  &  3.76 & 4.81 &22.45& 29.18\\ 
Unanswerable  & 79.51 &  78.48 &86.23& 85.45& -  &  88.96 & 88.94 &88.95& 93.89\\ 
 


 


\bottomrule
\end{tabular}
}
\caption{Performance of different answer types on the test set (zh). \model* represents \model~w/o Pre.}
\vspace{-5px}
\label{tab:answer_types}
\end{table*} 

%% file: acl-ijcnlp2021-templates/table/cheatable.tex
\begin{table}[t]
    \footnotesize
    \centering
    \resizebox{0.46\textwidth}{!}{
    \begin{tabular}{lccc}
    \toprule
    ~  & \bf RGNet & \bf w/o Passage  &  \bf w/o History \\
    \midrule
    Overall Performance & 62.94 & 19.31 & 52.67 \\
    $~~~~~~~~$Yes/No & 90.96 & 80.04 & 90.35 \\
    $~~~~~~~~$Comparison & 46.56 & 21.74 & 23.68 \\
    $~~~~~~~~$Arithmetic & 33.73 & 11.79 & 23.87\\
    $~~~~~~~~$Counterfactual & 29.18 & 11.22 & 22.89 \\
    $~~~~~~~~$Extract & 73.20 & 1.12 & 61.21 \\
    $~~~~~~~~$Unanswerable & 93.89 & 37.57 & 80.56 \\
    \midrule
    Reasoning Graph & 59.32 & 9.01 & 38.92 \\

    \bottomrule
    \end{tabular}}
\caption{Experiments w.r.t. \model{} w/ or w/o passages or historical QA pairs on our \dataset{} dataset.}
\vspace{-5px}
\label{tab:cheatable}
\end{table}

%% file: acl-ijcnlp2021-templates/table/transfer.tex
\begin{table}[h!]
\footnotesize
\centering
\resizebox{0.36\textwidth}{!}{
\begin{tabular}{l l | c  |  c }
\toprule

 Datasets & Models &   en & zh \\

\midrule

DropQA (en) &NumNet+ & 7.53 & -  \\\midrule
Math23K (zh)& GTS & - & 8.96   \\

\bottomrule
\end{tabular}}
\caption{Transfer results of other datasets on \dataset.}
\vspace{-10px}
\label{tab:transfer}
\end{table} 

%% file: acl-ijcnlp2021-templates/section/7_related_work.tex
\section{Related Work}

\noindent\textbf{Math Word Problems.}
In the past few years, there has been a growing number of datasets~\cite{wang-etal-2017-deep, miao-etal-2020-diverse,patel-etal-2021-nlp} and methods that have been proposed for MWPs, including statistical machine learning methods~\cite{mitra-baral-2016-learning,roy2018mapping}, semantic parsing methods~\cite{liang-etal-2018-meaning}, and deep learning methods~\cite{dns,mathdqn,seq2et,trnn,seq2tree,graph2tree,sau-solver,wu-etal-2020-knowledge}, emerging in the field of solving MWPs.

\noindent\textbf{Question Answering Datasets.}
This work mainly refers to conversational QA datasets~\cite{reddy2019coqa, choi2018quac, christmann2019look}, Multi-hop QA datasets~\cite{talmor2018web, yang2018hotpotqa, inoue2020r4c, ho2020constructing, chen2020hybridqa}, and discrete reasoning datasets~\cite{dua2019drop, sun2019dream}. For explanation in QA, CoQA~\cite{reddy2019coqa} provide rationale to make models understandable under conversations. HotpotQA~\cite{yang2018hotpotqa}, $\mathrm{R^4C}$~\cite{inoue2020r4c}, and 2WikiMultiHopQA~\cite{ho2020constructing}, provide a set of evidence to support training models to learn reasoning across paragraphs. In concurrent, ~\citet{dalvi2021entailmentbank} present a new dataset with explanations in the form of entailment trees, however, they do not consider numerical reasoning.

\noindent\textbf{Interpretable Reasoning.}
There have been many works~\cite{NEURIPS2020_fc84ad56, saha-etal-2020-prover, wolfson-etal-2020-break} exploring interpretable reasoning recently. Related to our work, \cite{dalvi2021entailmentbank} generates explanations in the form of entailment trees, namely a tree of entailment steps from known facts.

%% file: acl-ijcnlp2021-templates/section/8_conclusion.tex
\section{Conclusion}

In this work, we present a new QA datasets with complex numerical questions and interpretable reasoning graph. 
We also introduce an automatic evaluation metric for the generated reasoning process. We finally present an initial model producing the reasoning process while answering questions. The experiments show that \dataset{} is challenging and will become an interesting direction in both numerical QA and explainable QA.

\section{Ethics}
Finally, we want to state the cost and wage issues of our dataset annotation. We recruited 30 annotators from Chinese universities. We pay CNY1400 per 1000 samples, and it takes 83.33 hours to annotate every 1000 samples. Therefore, the hourly salary is about CNY16.8, equivalent to USD2.52 per hour. Please note that the minimum average hourly wage of the Sichuan Province of China (which is where the recruited annotators are from) is CNY16.3 per hour in 2018. Therefore, our pay is above the minimum average hourly wage. In total, We spent USD4500 in this dataset.

%% file: acl-ijcnlp2021-templates/9_appendix.tex
\pdfoutput=1






%
%


\renewcommand\thesection{\Alph{section}}

\setcounter{section}{0}
\section{Cross-Lingual Experiment}
\input{acl-ijcnlp2021-templates/table/multilingual}
In Table~\ref{tab:cross_lingual}, we conduct cross-lingual experiments on Chinese corpus and machine-translated English corpus. When training and testing in the same language, XLM-R pre-training achieved better results than normal RoBERTa; while in the case of monolingual training, the experimental effect of testing in another language was worse than that of testing in the same language. Finally, the best experimental results can be achieved by mixing the two languages corpus.
\section{Summary of Hyperparameters}
\input{acl-ijcnlp2021-templates/table/hyperparameters}
In Table~\ref{tab:hyperparameters}, We show the hyperparameters of the default model.

\section{Error Details}
In addition to grammatical and pragmatic errors in machine translation, which occur in traditional generic text domain, our dataset, due to the properties of MWPs and natural conversation, can generate some unique translation errors in machine translation, which deserve attention from cross-lingual research and are also the focus of manual error correction, as follows in Figure 1:

\begin{figure*}[ht]
\centering
\captionsetup{width=0.8\linewidth}
\includegraphics[width=0.8\textwidth]{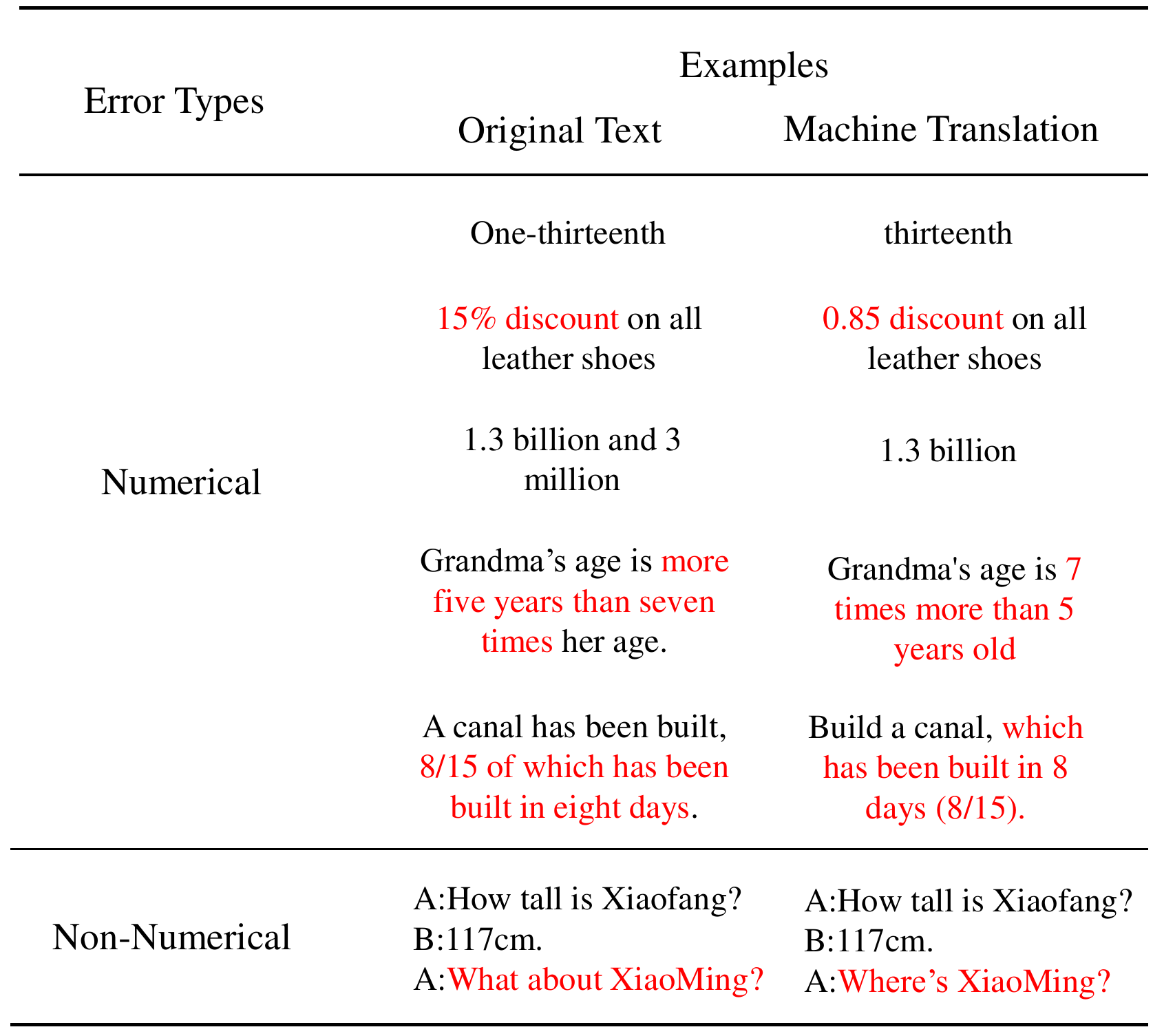}

\caption{Error Types and corresponding examples. "original text" represents that the original Chinese text is expressed in English accurately.}
\label{fig:app_eg1}
\end{figure*}

\section{Examples of Question-answer types}
\input{acl-ijcnlp2021-templates/table/distribution_example}
In Table~\ref{tab:qa_type_examples}, We show examples of different types of Question Answering.

\section{Detailed Examples}
We show two detailed examples in Figure~\ref{fig:app_eg1}, including passage, conversations, and reasoning graphs.
\begin{figure*}[ht]
\centering
\includegraphics[width=0.80\textwidth]{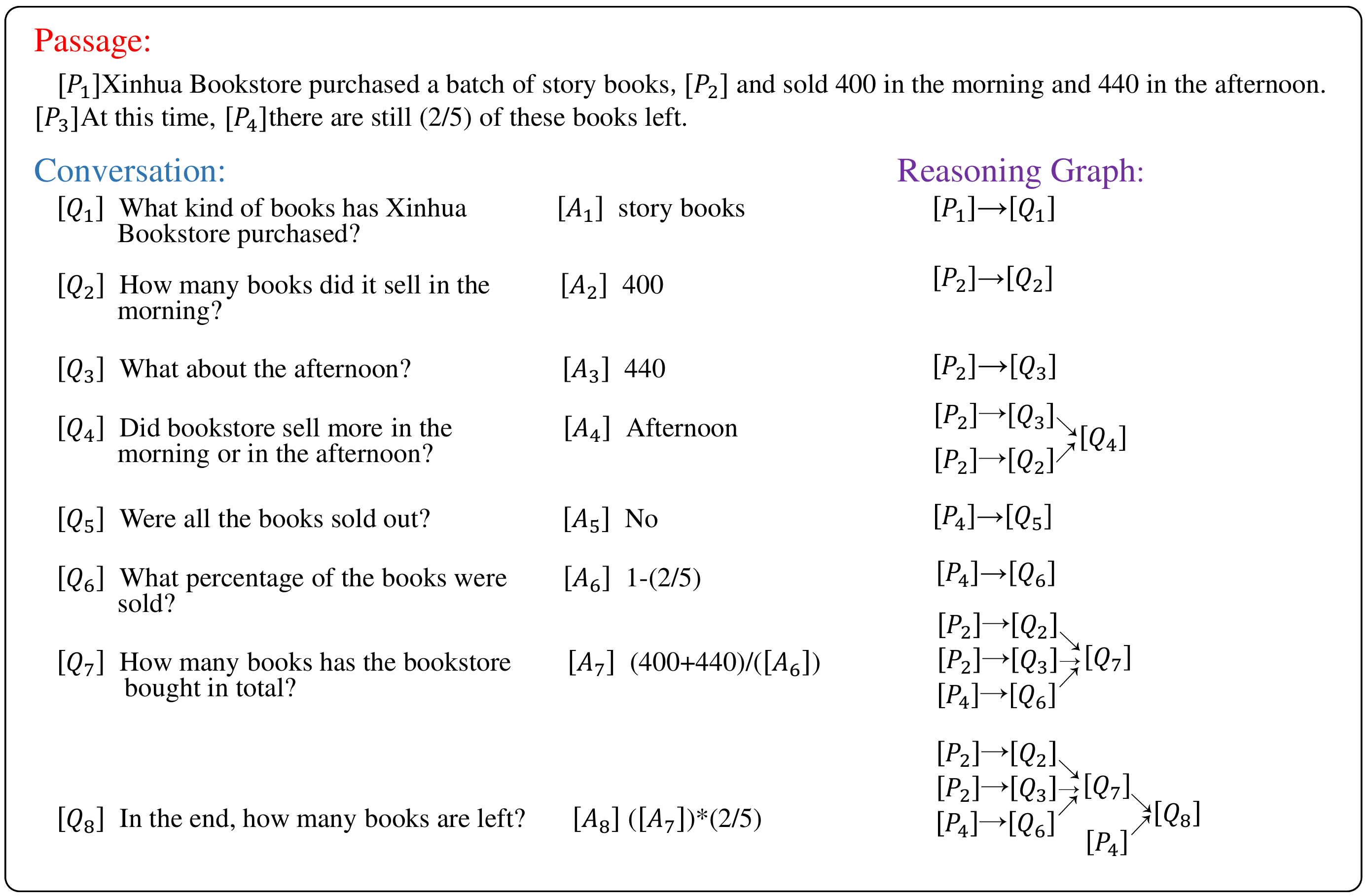}
\includegraphics[width=0.80\textwidth]{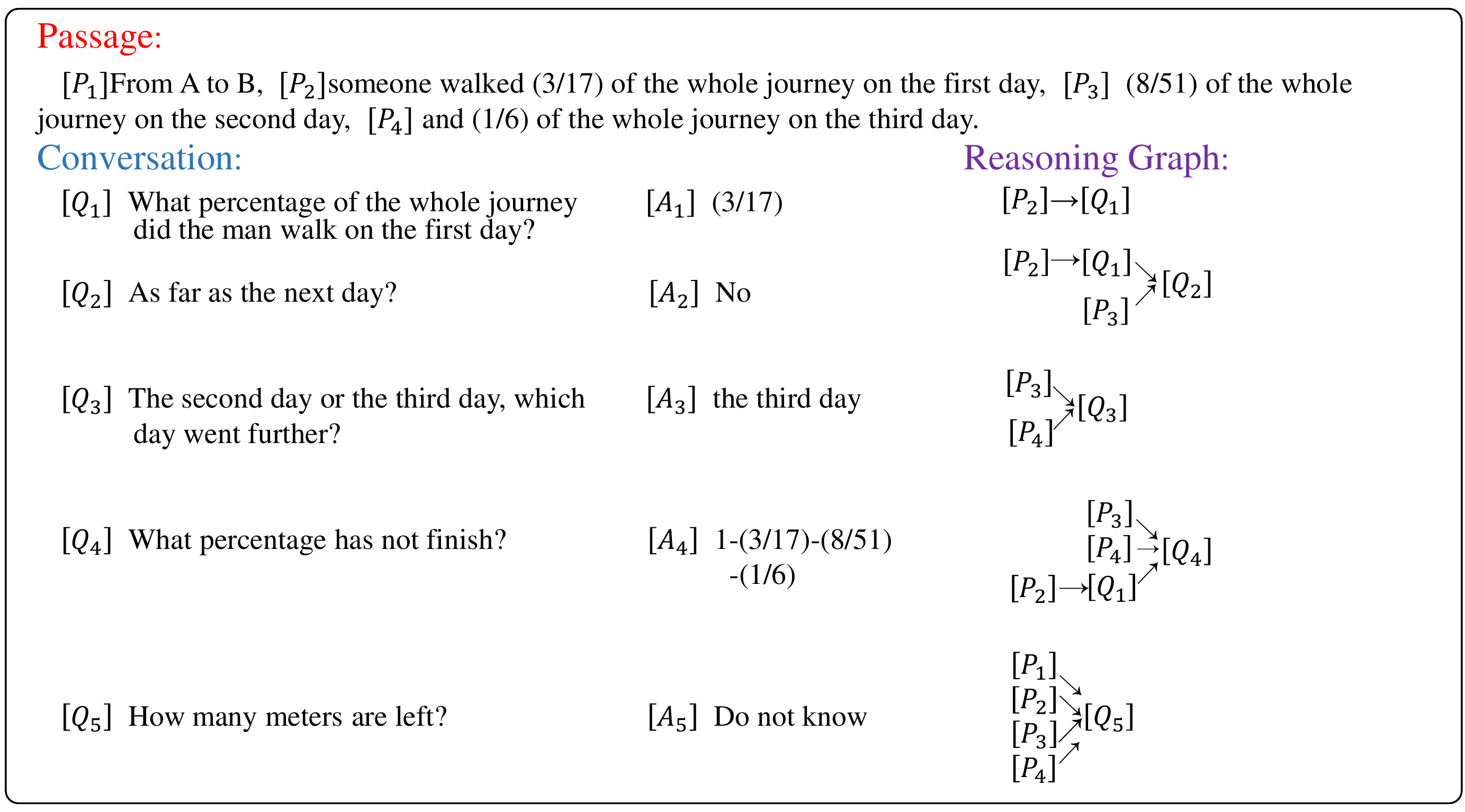}
\caption{Detailed Examples.} 
\label{fig:app_eg1}
\end{figure*}

\section{Interface of Conversation Collection}
\begin{figure*}[ht]
\centering
\includegraphics[width=0.96\textwidth]{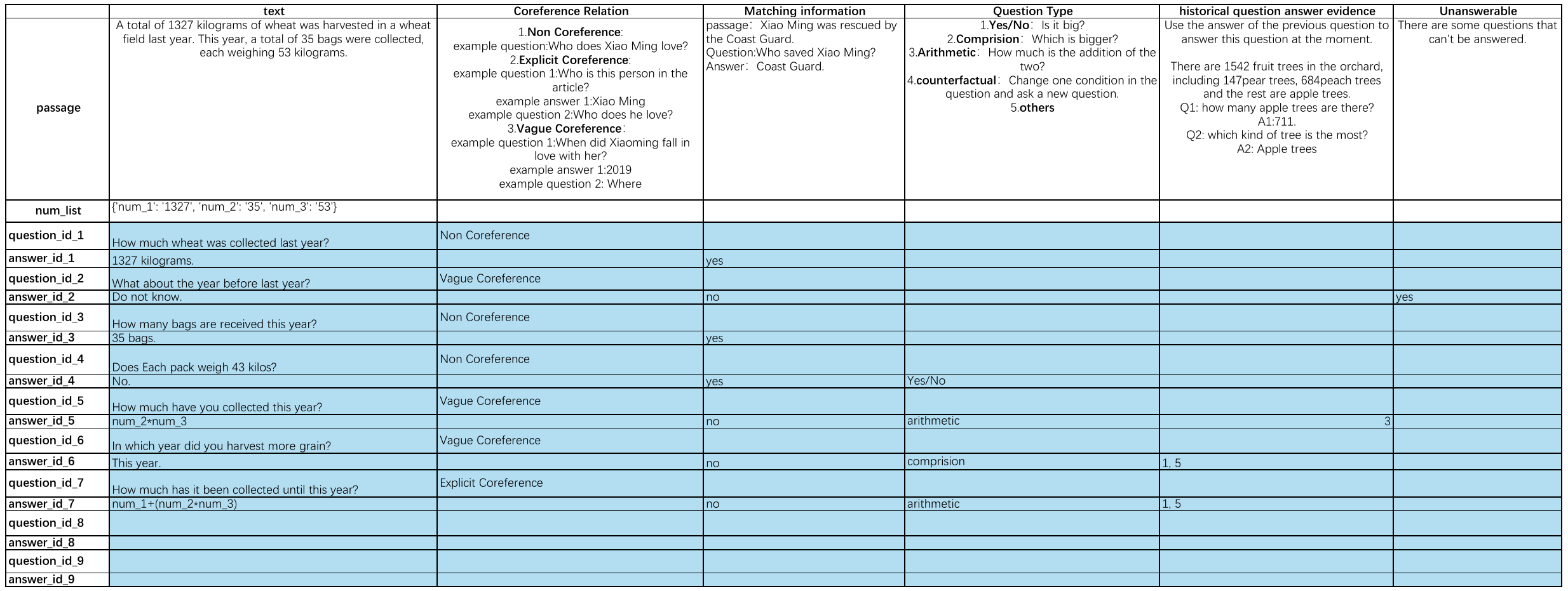}

\caption{Interface of Conversation Collection.} 
\label{fig:interface}
\end{figure*}

As depicted in Figure~\ref{fig:interface}, we show the translated interface for annotating the conversation about an passage. We automatically create an excel worksheet for each passage (the white part is automatically imported and filled in by the machine), and the annotator needs to follow the guidelines to complete the collection (the blue part is the part that the annotator needs to fill in). Column 2 is the area where the annotator completes the conversation collection, and the annotator fills in the desired conversation according to the content of the passage in (row 2, column 2) and the annotating guidelines. The annotator also needs to fill in some attributes of the corresponding question answer pair, column 3-7, respectively, coreference relationship (between the current question and historical question), Matching information (whether the answer directly corresponds to the phrase in the passage), Question Type, the historical question answer pairs evidence labeling, and unanswerable. After generating the conversation, the annotator needs to make a multi-choice among these attributes that corresponding to QA pair, meanwhile, We give explanation examples and definitions above.


%% file: acl-ijcnlp2021-templates/table/multilingual.tex
\begin{table}[!t]
\footnotesize
\centering
\begin{tabular}{l c | c c | c c }
\toprule
 \multirow{3}{*}{} & & \multicolumn{2}{c|}{en} & \multicolumn{2}{c}{zh} \\



Model & Train & Dev & Test & Dev & Test\\
\midrule
\multirow{2}{*}{RoBERTa} & en & 63.04 & 61.69 & -- & -- \\
& zh &-- & -- &  64.90  & 62.94 \\
\midrule
\multirow{3}{*}{XLM-R} & en & 64.42 & 62.26 & 57.00  & 54.13 \\
& zh & 48.63& 45.78& 68.84  & 66.76 \\
& en+zh & 66.37 & 63.89&  69.24 & 67.00 \\


\bottomrule
\end{tabular}
\caption{Performance comparison under different cross-lingual settings. 
}
\label{tab:cross_lingual}
\end{table} 

%% file: acl-ijcnlp2021-templates/table/hyperparameters.tex
\begin{table}[tb]
    \centering 
    \begin{tabular}{@{}lcc@{}}
    \toprule
         &  \model{} & \model{} w/o Pre \\
        \hline  
        train epoch  & 40 &  40  \\
        batch size  & 4 & 8 \\
        max length & 512 & 512 \\
        hidden size  & 768 & 128  \\
        num hidden layers  & 24 & --  \\
        num heads   & 24 & --  \\
        learning rate schedule  & -- & -- \\
        learning rate   & 5e-6 & 1e-4 \\
        dropout & 0.1 & 0.1  \\
        \bottomrule
    \end{tabular}
    \caption{Summary of hyperparameters derived from the defaults. Default hyperparameter sets are: \model{}, \model{} w/o Pre.}
    \label{tab:hyperparameters}
\end{table}

%% file: acl-ijcnlp2021-templates/table/distribution_example.tex
\begin{table*}[t]
    \centering
    \resizebox{0.98\textwidth}{!}{
    \begin{tabular}{lp{11.2cm}}
    \toprule
        Answer Type & Example  \\ \midrule
        
        Extraction  &P: The canteen has 580 kilograms of coal. It burns for 6 days. It burns 36 kilograms of coal every day. How many kilograms are left?\\
         & Q: How many kilograms does it burn per day? \\ 
         & A: 36 kilograms.  \\ \midrule

        Numerical Reasoning &  \\
        \,\,\,\,w/o external knowledge & Q: How much is the price of sandals?\\ 
         & A: $19$. ($10 + 10 \times 90\%.$) \\
         \,\,\,\,w/ external knowledge & Q: What is the circumference of the bottom surface?\\
        & A: $4.71$. ($\pi \times 1.5$) \\
        Counterfactual & Q: If the survival rate increased to $90\%$, how many saplings do you need?\\
        & A: $4666.67$. ($4200\div 90\%$) \\  \midrule
         
        Yes/No &Q: Is the BBK VCD price reduced?  \\
         & A: Yes. \\  \midrule
        
         Unanswerable  &Q: What brand of soy sauce is it?\\         
        & A: Do not know. \\  \midrule
        
        Comparison &Q: Which is more expensive, the original price or the current price?  \\
       & A: Original price. \\

    \bottomrule
    \end{tabular} }
    \caption{Examples of question-answer types in \dataset. }
    \label{tab:qa_type_examples}
\end{table*}